\documentclass{article}
\usepackage{ijcai20}

\usepackage{times}
\usepackage{soul}
\usepackage{url}
\usepackage[hidelinks]{hyperref}
\usepackage[utf8]{inputenc}
\usepackage[small]{caption}
\usepackage{graphicx}
\usepackage{amsmath}
\usepackage{amsthm}
\usepackage{booktabs}
\usepackage{algorithm}
\usepackage{algorithmic}
\usepackage{color} 
\usepackage{setspace}
\title{Complex Relation Extraction: Challenges and Opportunities}

\author{
Haiyun Jiang$^1$, 
Qiaoben Bao$^{1}$,
Qiao Cheng$^1$,
Deqing Yang$^2$,
Li Wang$^1$\And
Yanghua Xiao$^1$
\affiliations
   $^1$Shanghai Key Laboratory of Data Science, School of Computer Science, Fudan University, China \\
$^2$School of Data Science, Fudan University, Shanghai, China \\
\emails
}

\begin{document}
\maketitle

\begin{abstract}
 Relation extraction aims to identify the target relations of entities in texts.
 Relation extraction is very important for knowledge base construction and text understanding.
 Traditional binary relation extraction, including supervised, semi-supervised and distant supervised ones,  has been extensively studied and significant results are achieved.
In recent years, many complex relation extraction tasks, i.e., the variants of simple binary relation extraction, are proposed to meet the complex applications in practice.
However, there is no literature to fully investigate and summarize these complex relation extraction works so far.
In this paper, we first report the recent progress in traditional simple binary relation extraction.
Then we summarize the existing complex relation extraction tasks and  present the definition, recent progress, challenges and opportunities for each task.

\end{abstract}

\section{Introduction}
Relation extraction (RE) is one of the fundamental tasks in information extraction and it benefits a lot of natural language processing tasks, such as question answering, text understanding, etc.
RE is also a core step in the entire knowledge base construction pipeline.

Traditional RE tasks \cite{riedel2010modeling} aim to identify the correct relation between two entities from texts.
For example, we hope to extract the relational fact (\emph{Beijing, the-capital-of, China}) from the following text:
\begin{quote}
	\textit{\textcolor{red}{Beijing} formerly romanized as \textcolor{red}{Peking} is the capital of the People's Republic of \textcolor{red}{China}.} 
\end{quote}
Traditional RE tasks mainly focus on the binary relation between \emph{two} entities.
We refer to these tasks as binary relation extraction (BiRE for short) and they usually take learning-based solutions.
According to the problem settings, the traditional BiRE tasks are roughly divided into three categories: \emph{supervised BiRE, semi-supervised BiRE, distant supervised BiRE}.

Specifically, supervised BiRE focuses on the learning of a RE model from a set of \emph{high-quality} labeled data.
However, high-quality labeled data is difficult and costly to be obtained while unlabeled data is widely available.
Semi-supervised BiRE thus is proposed to train models with only a small set of labeled data and a large amount of unlabeled data.
Another effort to alleviate the difficulty of data labeling is distant supervision.
Distant supervised BiRE aims to learn a reliable RE model based on a set of weakly labeled samples.
The labels are obtained automatically in a heuristic way and usually contain a lot of noises.

Simple BiREs dominate the current research in information extraction.
In the early days, feature engineering and kernel-based methods are the focus of the research in supervised and distant supervised BiRE.
Bootstrapping was usually used in semi-supervised BiRE, where the relation instances and patterns are iteratively extracted based on a small set of seed instances.  
In recent years, with the development of deep learning, many advanced neural models, e.g., BERT, Transformer, capsule networks, are applied to RE tasks.

In  general,  simple BiRE has made significant progress and many effective solutions have been used in practice.
However, as intelligent applications fast grow, simple BiRE cannot meet the needs of these applications.
We elaborate the limitations of simple BiRE and introduce more complex RE tasks to solve these limitations.

\emph{First, simple BiRE depends on large amounts of data.}
However, it is difficult to obtain enough labeled (or unlabeled, noisy) data in many scenarios, which fails existing supervised (semi-supervised, distant supervised) RE models.
To solve this problem, the task of \textbf{\textit{few-shot relation extraction}} was proposed, which focuses on building effective models with just a few samples.
Some few-shot learning algorithms (e.g., metric learning-based \cite{Han2018FewRelAL}) are proven to be effective for this task.

\emph{Second, simple BiRE is limited to sentence-level extraction.}
Simple BiRE mainly focuses on the relation between an entity pair mentioned in a single sentence.
Instead, many other sources beyond sentences contain more rich semantic relation instances.
How to extract relations from various sources is an interesting and challenging problem.
Specifically,
\begin{itemize}
	\item  Many entity pairs appear in multiple sentences in a document, which cannot be extracted by simple BiRE models.
	 This motivates \textbf{\textit{Document-level extraction}}.
	
	\item Most works on BiRE only focus on monolingual (e.g, Chinese or English) corpus.
	But many entities are mentioned in multiple languages, indicating that it is possible to identify the relation using texts of different languages.
	\textbf{\textit{Cross-lingual relation extraction}} is thus proposed.
	
	\item In addition to texts, other modal (e.g., image, video) information is also useful for expressing certain semantic relations.
	For example, image is good at expressing the spatial relations.
	\textbf{\textit{Multi-modal relation extraction}} is proposed to use multi-modal information for RE.
\end{itemize}

\emph{Third, binary modeling in BiRE is far from satisfying the requirements of some complex applications.} 
The relationships of entities in the world are very complicated.
Binary relation in general is not enough to model the complicated semantics of real world and more complicated relation modeling is needed.
\begin{itemize}
	\item In some scenarios, we have to identify the relations involving multiple entities, i.e., \textbf{\textit{N-ary relation extraction}}.
	N-ary relation extraction aims to extract relations among $n$ entities in the context of one or more sentences.
	N-ary RE is very useful for document-level reading comprehension and supportive for question answering or document classification.
	
	\item Categorizing relations into different granularities is crucial for some tasks, such as building taxonomy, etc.
	\textbf{\textit{Multi-grained relation extraction}} aims to jointly extract multi-grained relations from texts.
	
	\item Many relational facts only hold true under certain conditions.
	\textbf{\textit{Conditional relation extraction}} aims to extract relations with certain constrains, e.g., temporal or spatial conditions, which are very important for complex applications. 
	For example, we know the fact (\emph{Obama, President, United States}) is only valid during 2008-2017.
	If this fact is used for knowledge-based question answering today, it may have serious political implications.
	
	\item Some facts can be expressed in a nested way.
	\textbf{\textit{Nested relation extraction}}  is proposed to extract this kind of knowledge.
\end{itemize}

\emph{Fourth, the existing BiRE cannot handle the overlapping entities well.}
For another example, the former sentence ``\textit{Beijing formerly romanized as Peking ...}'' contains three facts: (\emph{Beijing, the-capital-of, China}), (\emph{Beijing, the-same-as, Peking}) and  (\emph{Peking, the-capital-of, China}).
However, traditional BiRE tends to extract these facts independently, which losses much potential supervision information.
To model this property, the task of \textbf{\textit{overlapping RE}} is proposed, where one or two entities in two facts are overlapped.

In this paper, we refer to the RE tasks mentioned above as complex RE.
The contents of this paper contain two parts.
The first part (Sec 2) presents the summary of the traditional BiREs.
Besides, the challenges and directions in BiRE are also concluded.
The second part (Sec 3) introduces the complex RE tasks, including the definition, example and the recent progress.
Besides, we also present the research challenges and opportunities for these complex RE tasks.

We hope this survey will help researchers to understand the latest progress, challenges and opportunities of the sub-tasks in RE.


\section{Binary Relation Extraction}
Simple binary relation extraction (BiRE) has been extensively studied for many years.
In general, BiRE can be categorized into: supervised, semi-supervised, distant supervised paradigms.

\subsection{Supervised BiRE}
\textbf{Description.}
Supervised BiRE focuses on the learning of a RE model based on a set of \emph{high-quality} labeled samples.
These samples are widely obtained by manual annotation  or careful crowdsourcing.
Each sample is formalized as $(t,s_t,r)$, where $t=(e_1,e_2)$ is an entity pair.
$s_t$ is a sentence containing $t$ and it \emph{expresses} the labeled relation $r$.
A supervised BiRE model accepts $t,s_t$ as inputs and predicts the proper relation $r$ for entity pair $t$ as the output.


\textbf{Recent works.}
In recent years, deep learning has been extensively used in RE tasks and many novel neural models are proposed.
We highlight typical efforts in this direction.
\begin{itemize}
	\item  (1) Neural graph-based models. 
	Graph-based methods have been successfully applied to RE and obtain high performance. 
	For example, \cite{zhang2018graph} first applied graph convolutional network (GCN) to RE.
	
	\item  (2)  Pre-training based methods.
	Pre-trained models, e.g., BERT and XLNet,  can encode a given text into its proper distribution representation, i.e., text embedding.
	\cite{zhao2019improving} constructs entity pair graphs combined with the semantic features from BERT.
	
	\item  (3) Capsule network-based methods.
	For  example, \cite{zhang2019multi} takes capsule network with an attention-based routing algorithm to deal with the multi-label problem in RE. 
	
\end{itemize}

\textbf{SOAT results.}
The commonly used datasets in supervised BiRE include SemEval-2010 Task 8\footnote{https://mailman.uib.no/public/corpora/2010-February/010118.html},  ACE 2004\footnote{https://catalog.ldc.upenn.edu/LDC2005T09} and TACRED\footnote{https://nlp.stanford.edu/projects/tacred/}.
We present the state-of-the-art results of SemEval-2010 Task 8 in Table \ref{macro-f1}.
\begin{table}[htbp]
	\centering
	\scalebox{1}{
		\begin{tabular}{cccccc}
			\hline
			Model &Macro-F1 \\
			\hline
			TRE &87.1\\
			R-BERT &89.2 \\
			EPGNN &90.2\\
			\hline
		\end{tabular}
	}
	\caption{The SOAT results on SemEval-2010 Task 8. All the results are from [Zhao \emph{et al}., 2019].}
	\label{macro-f1}
\end{table}


\subsection{Semi-supervised BiRE}
\textbf{Description.}
In many scenarios, rich labeled data is difficult to be obtained, but a lot of unlabeled data is available.
To leverage the large amount of unlabeled data in the training stage, semi-supervised BiRE tries to learn from both labeled data and unlabeled data.

Formally speaking, we denote the pre-defined set of relations as $ \mathcal{R} $, a set of labeled data as  $ \mathcal{S}_{L} = \{(x_{i}, y_{i})\}_{i=1}^{N_{L}} $ and a set of unlabeled data as $ \mathcal{S}_{U} = \{x_{i}\}_{i=1}^{N_{U}} $, where $ y_{i} \in \mathcal{R} $ and $ N_{L} $ or $N_U$ is the corresponding data size.
Semi-supervised BiRE aims to learn a function $ F(\mathcal{S}_{L},\mathcal{S}_{U}, \mathcal{R}, x) $ that models both the labeled and unlabeled data and predicts the target relation $r$ for $x$.

\textbf{Recent works.}
As a main branch of semi-supervised BiRE, \emph{bootstrapping} starts from some labeled seed instances and learn a preliminary model which is used to find more labeled instances.
Many works also focus on alleviating semantic drift problem in bootstrapping.
For example, 
\cite{Carlson2010CoupledSL} adds constrains to the training procedure by  coupling many extractors for different categories and relations.

With the exploration of teacher-student models in semi-supervised learning,
\cite{Luo2019SemiSupervisedTA} introduces this architecture into semi-supervised BiRE where students learn a robust representation from unlabeled data and teachers guide students with labeled data.
Some other works also utilize multi-task learning by jointly learning semi-supervised BiRE task with other tasks.

\textbf{SOAT results.}
It is hard to fairly compare different semi-supervised BiRE models.
This is because: (1) Many bootstrapping approaches are deployed in an open world setting and extract relations on the web.
(2) Semi-supervised setting varies greatly between methods, i.e., 
the level of supervision, the data size of unlabeled data and the evaluation metrics can not be exactly the same in different methods.
For these reasons, we do not provide the SOTA here.
In general, semi-supervised BiRE has made great progress in recent years and many mature systems (e.g., DIPRE, Snowball and KnowItAll) have been applied to some structured knowledge acquisition tasks in practice.

\subsection{Distant Supervised BiRE}
\textbf{Description.}
Similar to supervised BiRE, each sample in distant supervised BiRE  can also be formalized as $(t,s_t,r)$.
The difference is that these samples are constructed in an automatic way, e.g., aligning knowledge base with text corpora \cite{riedel2010modeling}.
The strong assumption in sample acquisition makes the samples in distant supervised BiRE contain lots of noisy relation labels.
In other words, $s$ may weakly or not express the labeled relation $r$.
As a result, the main focus of research in distant supervised BiRE is how to alleviate the impact of noise on performance.

\textbf{Recent works.}
We highlight several popular directions in recent years.
\begin{itemize}
	\item The idea of reinforcement learning has been widely used in noise detection.
	For example, \cite{sun2019reinforcement} takes policy network to detect the noisy labels and further to obtain the latent correct labels.
	
	\item The adversarial training is also proven to be effective in improving the robustness of the RE model on the noisy samples \cite{liu2019adversarial}.
	
	\item Various attention mechanisms are also proposed to learn the important features or instances among the noisy samples \cite{li2019self}.
	Besides, some other techniques are also applied to noise detection, e.g., soft constraints of entity types,  variant CNN, Non-IID assumption, noise label converter and so on.
\end{itemize}

\textbf{SOAT results.}
The commonly used benchmark for distant supervised BiRE is NYT \cite{riedel2010modeling}, which is constructed by aligning triples in Freebase with texts in New York Times. 
We report the SOAT results in Table \ref{AUC}.
\begin{table}[htbp]
	\centering
	\scalebox{1}{
		\begin{tabular}{cccccc}
			\hline
			Model &AUC \\
			\hline
PCNN+HATT &0.42 \\
PCNN+ATT-RA+BAG-ATT& 0.42\\
SeG & 0.51\\
			\hline
		\end{tabular}
	}
	\caption{The SOAT results on NYT, where all the results are from [Li \emph{et al}., 2019].  AUC denotes the area under the precision-recall curve.}
	\label{AUC}
\end{table}

\subsection{ Challenges and Directions of BiRE}
Although simple BiRE has made great progress in recent years, there are still some challenges.
\begin{itemize}
	\item  \textbf{\textit{Reliability of benchmarks}}. 
	A reliable benchmark can be measured from two aspects: scale and quality.
	That is, a good benchmark should contain large-scale and high-quality test samples.
	However, the two conditions cannot be easy simultaneously satisfied in BiRE tasks.
	For example, in supervised BiRE, the scale of the test set is usually very small.
	How to obtain reliable benchmarks is a promising direction.
	
	\item  \textbf{\textit{Reliability of model learning}}.
	 Because of the various factors, e.g., limited data or noise, the precise semantic features of relations are still difficult to be captured.
	 With the development of machine learning technology, e.g., pre-training, transfer learning, this problem can be alleviated to some extent.
	 But learning a highly reliable BiRE model is still an important direction.
	 
	\item  \textbf{\textit{Sparse detection in applications}}. 
	In real applications, we usually pre-define a set of relations $\mathcal{R} $ and then extract their instances from massive candidate entity pairs.
	However, there are countless semantic relations in the world and most of the candidate entity pairs to be processed have no relation in $\mathcal{R}$.
	How to detect the correct instances from a huge collection is a big challenge and we refer to it as \emph{sparse detection.}
	Fortunately, there are some benchmarks to model this property.
	For example, NYT contains relation ``NA'' that  denotes there is no relation in the relation set.
	But to the best of our knowledge, there is no RE model that performs very well on sparse detection.

\end{itemize}


\section{Complex Relation Extraction}
There are only very recent works around Complex Relation Extraction (CoRE).
Different from conventional BiRE, CoRE tries to extract more complex relations that involve multiple entities or under certain constrains.
In this section,  we present the definition and investigate the recent progress for each complex RE task.
We also conclude the challenges of each task.

\subsection{Few-shot Relation Extraction}
In most cases, a relation only has fewer instances, which makes the traditional supervised RE models powerless.
As a new paradigm, few-shot learning tends to be effective for this problem, i.e., few-shot RE.

Few-shot RE can be formalized as follows \cite{Han2018FewRelAL}.
Given a set of relations $\mathcal R$ with $n$ relations, a supporting set $\mathcal S$ is denoted as:
\begin {equation} 
\begin{split}
	\mathcal S = &\{(x_1^1,r_1),(x_2^1,r_1),...,(x_{m_1}^1,r_1), \\
	& ..., \\
	&(x_1^n,r_n),(x_2^n,r_n),...,(x_{m_n}^n,r_n)\} \\
\end{split}
\end {equation}
where $r_i \in \mathcal R$ is the $i$-th relation.
$x_j^i$ is a sentence with an entity pair that is labeled with relation $r_i$.
Few-shot RE aims to learn a function $F(\mathcal S, \mathcal R, x)$ and predicts the proper relation $y$ for the unlabeled sample $x$.

\cite{Han2018FewRelAL} proposes a new few-shot RE dataset: FewRel.
They also implemented the recent SOAT few-shot learning algorithms on this dataset.
FewRel 2.0 \cite{Gao2019FewRel2T} is a more challenging few-shot dataset, which aims to study the problem of new domain extraction under few instances.
Considering the noise in few-shot RE, \cite{Gao2019HybridAP} proposes a hybrid attention-based prototypical network to extract informative features.
\cite{Soares2019MatchingTB} takes BERT to learn the distributional similarity between two sentences where the entity pair in the sentence is replaced by  a [BLANK] symbol.


There are two challenges or directions for few-shot RE:
\begin{itemize}
	\item  \textbf{\textit{Relative importance of samples}}:
	Since there is only a few instances for each relation, it is very necessary to learn from other relation instances when learning semantic features of a target relation.
	As a result, the relative importance of an instance to the target relation should be learned, otherwise, the noise instance will be introduced.
 
	\item  \textbf{\textit{Knowledge reasoning is needed}}:
	As \cite{Han2018FewRelAL} points out, the relation prediction in a large number of samples need deep reasoning beyond the text in the sample.
	For example, given  the sentence \cite{Han2018FewRelAL}:
	\begin{quote}
		\textit{He was a professor at \textcolor{red}{Reed College}, where he taught \textcolor{red}{Steve Jobs}, and replaced Lloyd J. Reynolds as the head of the calligraphy program.} 
	\end{quote}
The logical reasoning with common sense is needed to infer the relational fact (\emph{Steve Jobs, educated-at, Reed College}).
\end{itemize}

\subsection{Document Relation Extraction} 
Document relation extraction aims to extract relations between entity mentions at document-level.
In this task, the relation mentions can span multiple sentences and even paragraphs.
These properties make the problem more challenging compared to intra-sentence relation extraction.

In the real world case of RE, the data to be processed is often in document form which requires RE model of document-level, or inter-sentence rather than intra-sentence.
Until recently, methods like \cite{Song2018NaryRE,Sahu2019IntersentenceRE} appear and accelerate the development of inter-sentence RE tasks. All of these methods can be used to enrich the knowledge base.

\cite{Peng2017CrossSentenceNR} first presents graph-based LSTM model to solve RE in multiple sentences.
They build two directed acyclic graphs based on the word dependencies and then utilize LSTM to learn the hidden presentations. \cite{Song2018NaryRE} further improves the model and introduce a graph-state LSTM model which can keep whole graph information and has high efficacy in training and decoding steps.
\cite{Sahu2019IntersentenceRE} uses five type edges to build the document-level graph and learn presentation by a labeled edge GCNN model.
Bi-affine layer aggregates all entity mentions and generates the final relation prediction.

The main difficulties in document RE are as follows:
\begin{itemize}
	\item  \textbf{\textit{Diversity of document format}}: The document can be in a variety of formats. The first task is to transform the original data file like \emph{pdf} to the specific format like \textit{txt}.
	Some important information can be lost in this step and make it harder to conduct RE process.
	
	\item  \textbf{\textit{Long dependencies cross sentences}}: Relation mentions can span long distance in the document.
	Traditional CNN and RNN based networks fail to capture those features in longer sequences.
	
\end{itemize}

\subsection{Cross-lingual Relation  Extraction}
Cross-lingual RE seeks to learn an extractor trained in the resource-rich language and transfers it to the target language.
The cross-lingual RE also takes the sentence as well as entity mentions as input and outputs the relation between the given entity pair.

Cross-lingual RE is beneficial to the knowledge base completion because some entities may appear more frequently in corpus of a certain language.
Considering the lack of well-annotated data, it will inevitably lose a lot of informative facts that could not be extracted based on the resource-poor language.
Cross-lingual RE can address this drawback exactly.

At an early age, cross-lingual RE methods depend on parallel corpora and then conduct extraction by projecting the source language to the target one.
\cite{Zou2018AdversarialFA} solves the problems with the help of translation mechanism.
When cross-lingual word embedding was proposed, \cite{Ni2019NeuralCR} utilizes it to map source embeddings to the target.
\cite{Subburathinam2019CrosslingualST} first builds text features by universal dependency parsing tools and then adopts GCN to learn the hidden presentations in the shared semantic space in which the RE of both languages can be conducted.

Although there are many methods of cross-lingual RE, main challenges are as follows:
\begin{itemize}
	\item   \textbf{\textit{Gaps between languages}}: The gaps between languages are quite different.
	Whether some shared features are useful universally is still a question.
	
	\item   \textbf{\textit{Applicable in practice}}: The proposed state-of-the-art models do not achieve the satisfactory outcome that the best performance of F1-score is around 62\%.
	Due to the relatively low performance, it is not reliable to use in practice.
	
\end{itemize}

\subsection{Multi-modal Relation Extraction}
With the explosive growth of information from Internet, images and videos also become rich resource.
Multi-modal RE take the advantage of these large scale corpus and focus on extracting relations from them.

As a vivid way to convey information, images and videos can implicate much knowledge.
On the one hand, humans are like to express some common sense knowledge using images rather than explicitly saying it.
On the other hand, combining multi-modal corpus has shown promising results in many tasks.
This phenomena highlights the importance of harvesting relations from images or videos rather than just free text.

\cite{Chen2013NEILEV} introduced a Never Ending Image Learner (NEIL) for visual knowledge from Internet.
NEIL cluster images from websites and mining relationships between instances.
Based on recently proposed visual question answering (VQA) datasets, multi-modal RE like \emph{Object-Object} or \emph{Object-Attribute} relations also aroused wide concern.

Multi-modal RE is still a research hotspot and have many interesting problems to focus on:
\begin{itemize}
	\item  \textbf{\textit{Generality of knowledge}}:
	Relation extracted by existing works is highly related to given input, e.g., ``the ball in the picture is red''.
	These knowledge is shallow and hard to be utilized in further study.
	How to extract connotative but informative relation, e.g., common sense knowledge, is one of the ultimate goals of multi-modal RE.
	\item  \textbf{\textit{Multi-modal KB}}:
	Although researches have begun to construct multi-modal KBs, there is still a big gap between it and existing KBs.
\end{itemize}

\subsection{N-ary Relation Extraction}
N-ary relation extraction (NRE) aims to extract relations among $n$ entities in the context of one or more sentences.
In the NRE task, the input can be denoted as $(E, T)$, where $E = (e_1,e_2,…,e_n)$ contains all the entity mentions and $T = (s_1,s_2,…,s_m)$ is the given text containing $m$ sentences ($m\ge1$). 
The target is to predict the relation among those $n$ entities.
The relation set is pre-defined and represented as $R = (r_1,r_2,…,r_k)$, where ``NA''  is also included in $R$, denoting there is no relation among the $n$ entities.

For example, given the text:
\begin{quote}
	\textit{The deletion mutation on exon-19 of \textcolor{red}{EGFR} gene was present in 16 patients, while the \textcolor{red}{858E} point mutation on exon-21 was noted in 10. All patients were treated with \textcolor{red}{gefitinib} and showed a partial response.} \cite{Peng2017CrossSentenceNR}
\end{quote}
the entity mentions (\emph{gefitinib, EGFR, 858E}) form  the relation of (\emph{drug, gene, mutation}).

N-ary RE has attracted more research interests.
It plays a crucial role in applications such as detecting cause-effect and predicting drug-gene-mutation facts. 
In contrast to the prosperity of the binary RE, there are fewer methods proposed for N-ary RE task.
\cite{McDonald2005SimpleAF} studies the case in biomedical domain where the n-ary relation is expressed within a single sentence.
Recently, some exciting work \cite{Song2018NaryRE}  has appeared in the task of named entity recognition. 
These methods can well handle the cross-sentence N-ary RE.

\cite{Peng2017CrossSentenceNR} explores a general RE framework based on graph LSTM.
They first transform the text input to graph by regarding the words as nodes and the dependencies, links between adjacent words and inter-sentence relations as edges.
Secondly, the graph is separated into two DAGs from which they utilize an extended tree LSTM network to learn the text presentations in the third step.
\cite{Song2018NaryRE} propose a graph-state LSTM model that can learn better presentation of the input text during the recurrent graph state transition.
With increasing numbers of recurrent steps, each word can capture more information from a larger context.

There is still some suffering in N-ary RE.
\begin{itemize}
	\item  \textbf{\textit{Lack of data}}: There is no well-annotated data for evaluating N-ary RE specially. 
	The currently used benchmark dataset in \cite{Peng2017CrossSentenceNR} and \cite{Song2018NaryRE} is constructed based on distant supervision and has only one ternary relation.
	
	\item  \textbf{\textit{Absence of end-to-end model}}: Most of the preliminary works require all the entity mentions as input but obtaining them can be demanding. 
	In real applications, an end-to-end model can make a vast difference.
\end{itemize}

\subsection{Multi-grained Relation Extraction}
Multi-grained relation describes knowledge in a coarse-to-fine manner.
Intuitively, fine-grained relation aims to distinguish subordinate-level relations from a coarse-grained relation.
Categorizing relations into different levels is crucial for building taxonomy, mining level-specific information, etc.

The performance gains from multi-grained information have been demonstrated in many tasks.
Multi-grained relations via categories can provide additional supervision from annotated data~\cite{Wu2019GlyceGF}.
By learning multi-grained topics of documents, \cite{Chen2011ShortTC} further improves the performance of short text classification.


However, multi-grained RE attracted little attention from the community recently.
\cite{Xia2019MultiGrainedNE} proposes a multi-grained named entity recognition framework to tackle the overlapping and nested problem.

There are several possible reasons for the slow progress of multi-grained RE:
\begin{itemize}
	\item  \textbf{\textit{Fuzzy boundary}}:
	The boundary between coarse-grained and fine-grained relations is not clear.
	Existing works almost regard two type relations as a whole part, which leads to a bottleneck of performance.
	\item  \textbf{\textit{Fairness of evaluation metrics}}:
	The naive F1 measurement is not sufficient for multi-grained RE, and  corresponding evaluation metrics are needed for better reflect the quality in a multi-grained manner.
\end{itemize}

\subsection{Conditional Relation Extraction}
Conditional RE aims to extract relations with certain constrains, e.g., temporal or spatial conditions.
A conditional relation can be generally denoted as $ (s, p, o, c) $, where $ (s, p, o) $ is the original \textit{subject–property–object} triple and $ c $ is the condition to hold relation true.
Take temporal condition as an example, the relation \textit{PresidentOf}(Barack Hussein Obama, American) only holds true over the temporal period 2008 to 2017.
So the condition $ c $ can be a temporal interval $ [2008, 2017] $ here.

Nowadays, large scale KB contains millions of entity and relation instances, such as DBpedia, Freebase and YAGO.
However, few of them consider relation as conditioned or include the above mentioned external condition in the KB.
It heavily limits the applicability of existing KB to sophisticated reasoning task and leads to an urgent need for research in conditional RE.

In the early stage, pattern based methods combined with manually designed features are adopted to capture the condition.
YAGO2 uses regular expressions to extract temporal and spatial relations from Wikipedia infobox.
\cite{Liu2019ExtractingTP} try to generate patterns for time-variant relations.
Some machine learning approaches are also tried in medical field~\cite{Gurulingappa2012ExtractionOP}.

Preliminary works have been attempted, but there is still no good solution.
The main challenges are as follows:
\begin{itemize}
	\item  \textbf{\textit{Complexity of dependencies}}:
	The complex dependencies between entities, relation and its condition make it hard to handle different part properly.
	
	\item  \textbf{\textit{Flexibility of condition}}:
	Condition in free text may have many existing forms. A general framework is needed to formalize the conditional dimension.
	
	\item  \textbf{\textit{Lack of data}}:
	There is no well-annotated data for conditional RE yet. To some extent, it prevents using end-to-end model for this task.
	
\end{itemize}

\subsection{Nested Relation Extraction}
Traditional BiRE can be expressed as (\emph{arg1, rel, arg2}), while nested RE can be formalized as (\emph{arg1, rel,} (\emph{arg2, rel2, arg3})) or ((\emph{arg1, rel, arg2}), \emph{rel2, arg3}).
However, traditional binary RE will lose some information, resulting in incomplete and uninformative triples. 
Nested RE helps to express the meaning of the original sentence more correctly. 
In addition, nested RE will be more beneficial to downstream tasks, such as question answering that relies heavily on the correctness and completeness of triples.

Some recent works try to study nested RE. 
NESTIE \cite{bhutani2016nested} learns syntactic patterns for the relations that are expressed as nested templates.
StuffIE \cite{prasojo2018stuffie} exploits Stanford dependency parsing and lexical databases to extract nested relations.

However, the existing works are not good enough, mainly because of the following challenges:
\begin{itemize}
	\item  \textbf{\textit{Complexity of structure}} Sentences consist of many clauses, or there are many entities and nested relationships. Sentence structure is often very complex and it is difficult to analyze nested structures directly. 
	\item  \textbf{\textit{Implicit subject}} In a sentence, the subject may appear only once, most of which are in the form of a reference (him, he), and the corresponding real entity needs to be found.
\end{itemize}

\subsection{Overlapping Relation Extraction}
Different relational triples in a sentence may have different degree overlap. 
\cite{zeng2018extracting} concludes two overlapping types: Entity Pair Overlap (EPO) and Single Entity Overlap (SEO). 
EPO means some triplets share overlapping entity pairs. 
SEO means some triplets share an overlapped entity but they don’t share overlapped entity pair. 
For example, ($s_1$, \emph{president of}, $o_1$) and ($s_1$, \emph{born in}, $o_2$) are SEO, which share the same entity ``s1". 
($s_1$, \emph{born in}, $o_2$) and ($s_1$, \emph{live in}, $o_2$) belong to EPO, which share the same entity pair ``($s_1$,$o_2$)".

The previous RE was designed to find relation based on given entity pairs.
However, in actual applications, the location of entities is often unknown, and there may be multiple relationships between entities. Ignoring overlapping RE will lose many related triples, leading to incomplete knowledge graphs.

Some researchers are now studying how to consider overlapping relations in a sentence. CopyR \cite{zeng2018extracting} adopts an end-to-end neural model  with copy mechanism to extract overlapping relations, which is the first work considering overlap problems. \cite{takanobu2019hierarchical} designs a hierarchical paradigm, which incorporates reinforcement learning to extract overlapping relations.

But there are still some challenges in overlapping relationship extraction.
\begin{itemize}
	\item  \textbf{\textit{Complexity of relations}}: There may be no relationship or multiple relationships between two entities in a sentence.
	\item  \textbf{\textit{Unknown entities and relationships}}: The locations of entities and relationships are unknown, and it is difficult to find them correctly.
\end{itemize}

\section{Conclusion}
Relation extraction denotes a series of tasks that aim to identify the proper relations for two or more entities under  specific settings.
In this paper,  we summarize the latest progress of simple binary RE tasks, including supervised, semi-supervised and distant supervised RE.
Furthermore, we also investigate the more complex RE tasks, including the definition, recent progress, challenges and opportunities.

Relationship extraction research is an eternal proposition, which mainly benefits from the advances in natural language processing and machine learning.
We argue that the existing progress made in simple BiRE can also be transferred to the complex RE tasks.
Mature applications for complex RE are still far away.
We hope this survey make researchers quickly understand the concepts and research progress of each subtask in complex RE.



\clearpage

\bibliographystyle{named}
\bibliography{ijcai20}
\end{document}